\documentclass[letterpaper]{article}

\usepackage{hyperref}            
\usepackage{amsfonts}       
\usepackage{nicefrac}       
\usepackage{color}
\usepackage{amssymb} 
\usepackage{amsmath} 
\usepackage{dsfont}
\usepackage{soul}
\usepackage{subcaption}
\usepackage{bbm}
\usepackage{enumitem}

\usepackage{tikz}
\usetikzlibrary{calc,intersections,through,backgrounds}
\usepackage[latin1]{inputenc}

\usepackage{caption}
\usepackage{subcaption}
\usepackage{graphicx}
\usepackage{graphics}

\usepackage[linesnumbered,ruled,vlined]{algorithm2e}

\newcommand{\RR}{\mathbb{R}}

\begin{document}

\title{May the force be with you}
\author{Yulan Zhang$^{1}$ and Anna C.~Gilbert$^{2}$ and Stefan Steinerberger$^{3}$
\thanks{$^{1}$ This work was performed while Yulan Zhang was an undergraduate student at Yale University,
        {\tt\small yulan.zhang@yale.edu}}%
\thanks{$^{2}$Anna C.~Gilbert is with the Department of Mathematics,
         Yale University, New Haven, CT
        {\tt\small anna.gilbert@yale.edu}}%
\thanks{$^{3}$Stefan Steinerberger is with the Department of Mathematics,
        University of Washington, Seattle, WA
        {\tt\small steinerb@uw.edu}}%
}

\maketitle

\begin{abstract}
Modern methods in dimensionality reduction are dominated by nonlinear attraction-repulsion force-based methods (this includes t-SNE, UMAP, ForceAtlas2, LargeVis, and many more). The purpose of this paper is to demonstrate that all such methods, by design, come with an additional feature that is being automatically computed along the way, namely the vector field associated with these forces. We show how this vector field gives additional high-quality information and propose a general refinement strategy based on ideas from Morse theory. The efficiency of these ideas is illustrated specifically using t-SNE on synthetic and real-life data sets.
\end{abstract}


\section{Introduction}


Dimensionality reduction has become one of the fundamental problems in modern mathematical data science.
Standard methods for embedding data or dimension reduction include t-SNE~\cite{lvdm08}, PCA~\cite{shlens}, UMAP~\cite{mcinnes} and many others. There is an interesting and important gap between methods with solid theoretical footing (e.g., PCA and SVD, multi-dimensional scaling, and spectral methods) and methods which are less well understood albeit widely used in practice (e.g., t-SNE or UMAP).

We focus on t-SNE (which is closely related to UMAP ~\cite{bohm,vsumap}) because it is widely used for data visualization, dimensionality reduction, and clustering, yet it lacks extensive theoretical analysis. Because this method is so effective for visualizing well-clustered, high-dimensional data, it has become an indispensable tool for biological data analysis tasks, including, but not limited to single-cell transcriptomics ~\cite{transcriptomics}, single-cell flow, mass cytometry ~\cite{amir,unen}, and whole-genome sequencing ~\cite{li,diaz}. Sometimes directly and sometimes indirectly, scientists devise new experiments, draw scientific conclusions, and interpret experimental data based results from applying t-SNE to their data sets. It is a remarkable fact that such central tools in the life sciences are so poorly understood in their theoretical foundations.

The first theoretical result for the behavior of t-SNE was obtained by Linderman and Steinerberger~\cite{linderman} who proved that highly clustered data leads to a clustered output; this result has been refined and extended by Arora, et al.~\cite{arora} and by Cai and Ma~\cite{cai}. These theoretical guarantees, naturally, are based on assumptions that are perhaps not always met in practice; they do, however, offer a glimpse into the underlying mechanism of the method. 

Practitioners have argued that theory is not really necessary: Laurens van der Maaten famously replies in his FAQ to the question, `Why doesn't t-SNE work as well as LLE or Isomap on the Swiss roll data?' with `$\ldots$who cares about Swiss rolls when you can embed complex real-world data nicely?' \cite{laurens}. On the contrary, some practitioners claim that t-SNE figures can be manipulated to reveal almost anything (according to \cite{ChariBenerjeePachter} even an elephant!).

From all of this disagreement, we conclude that such a widely used algorithm is still not well-understood and its analysis requires additional work. The rate at which researchers propose new algorithms in machine learning and data science often outpaces the rate at which we develop theory to explain them. At the same time, a careful analysis of any individual algorithm may well reveal underlying principles that extend beyond the algorithm itself. This is the approach we take in this paper.

\begin{figure}[h!]
\centering
\includegraphics[width=4cm]{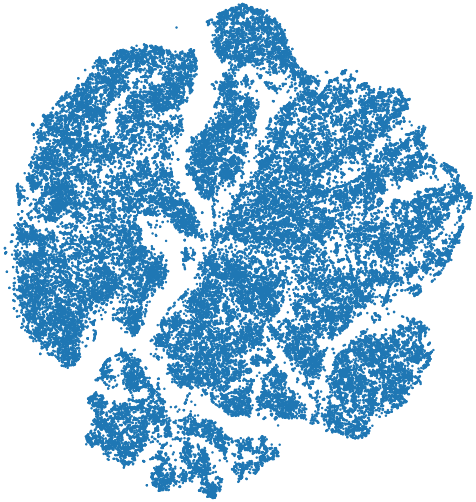}
\includegraphics[width=4cm]{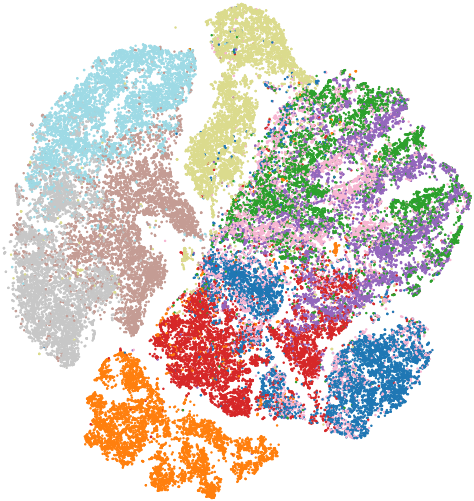}
\vspace{-5pt}
\caption{Standard t-SNE output for Fashion-MNIST \cite{fashion} dataset without (left) and with (right) ground truth labels. The ``correct'' clustering of the left embedding is ambiguous, and many apparent clusters mix the ground truth classes. Can we analyze the internal structure of a t-SNE cluster to recover this information? In the exploratory setting where t-SNE is typically used, this would be helpful for interpreting embeddings when no ground-truth is known.}
\label{fig:vector_flow:gauss:vectors}
\end{figure}
We begin with the empirical observation that although these methods are great at generating distinct clusters in the embedding, the global organization of these clusters is not always meaningful. In some sense, the methods are somewhat overeager when it comes to generating clusters. Both t-SNE and UMAP famously fail when the input data are from a continuous manifold: the manifold is broken up into random sub-clusters that cannot be reasonably said to be present in the original data. To make matters more complicated, given a particular output by one of these methods, we end up with sets of points in $\mathbb{R}^2$ which then require interpretation (again, a key effort in much of the life sciences). If the original data are already highly clustered, then the final clusters in $\mathbb{R}^2$ tend to be fairly pronounced and easy to interpret. If the original data are not distinctly clustered, then these point sets tend to be more diffuse and move into each other; as a result, interpretation requires additional considerable care\footnote{We note that separating embedded clusters is a problem independent of which embedding method is being used.}.

The main contribution of this paper is an important step towards reconciliation between theory and practice and clarification of interpretation of t-SNE and related methods. One may view t-SNE and a host of other embedding algorithms as placing points in $\RR^2$ so as to minimize a cost function, the gradient of which can be decomposed into attractive and repulsive forces which act upon the points and which determine the configuration of the points~\cite{linderman,arora}. We exploit an additional feature that is (a) \textit{already built into all attraction-repulsion based methods} (t-SNE, UMAP, ForceAtlas2, LargeVis, Laplacian eigenmaps, and many more), (b) helps in separating the boundary of embedded clusters as well as in elucidating the internal structure of any individual cluster, and (c) has been overlooked. This feature, \textit{which is already computed when running these methods}, provides additional information that is, in the current implementations, being thrown away. For simplicity of exposition, we will use t-SNE as a consistent example but the idea explained in this paper are equally applicable to all attraction-repulsion methods. The method of Laplacian eigenmaps is a particular outlier; because it is linear, our idea can be analyzed in closed form for that method which we do in Section~\ref{sec:spectral}.

Our idea may roughly be summarized as follows: start by running the method until one has reached an equilibrium (or near-equilibrium). Typically, it is not the case that the forces acting on any individual point have vanished---merely the effective force has vanished
because the attractive force (pulling a point towards similar points) and the repulsive force (moving the point away from other points) cancel each other (or very nearly cancel each other). Thus, $\mbox{[attractive force]} + \mbox{[repulsive force]} \sim 0$ and these two vectors will point in very nearly opposite directions but generically neither of these vectors will vanish. 

Our main insight is that \textbf{either one of these vectors can be used to provide additional features.} Indeed, points that end up close in the embedding but are rather different will have very different affinities and will be pulled in different directions which thus helps with disambiguation of cluster boundaries. The attractive vector was first proposed as an object of interest by Zhang and Steinerberger~\cite{zhang}. We show that we have an actionable algorithm that exploits the Morse-theoretic properties of an induced vector field to (a) separate t-SNE clusters and (b) uncover their internal structure (c) using only information that has been computed as part of the embedding process. That is, we examine the partition of $\RR^2$ based upon the vector field; we consider the sinks and the flow lines along the vector field towards these sinks.

We begin with the detailed definition of our algorithm in Section~\ref{sec:background} and conclude with a thorough empirical analysis of the algorithm on both illustrative and explorative examples in Section~\ref{sec:experiments}.

\section{Background}
\label{sec:background}

Dimensionality reduction techniques aim to embed a high dimensional input set $\mathcal{X} = \{x_1, \ldots, x_n\} \in \RR^N$ into a lower dimensional output space. For data visualization, this is typically $\RR^2$ or $\RR^3$. We denote the output embedding as $\mathcal{Y} = \{y_1, \ldots, y_n\}$. These methods usually aim to preserve some type of structure in the original data. For example, PCA is a linear method that projects the input along the principal axes where it has maximum variance ~\cite{shlens}. In other words, it tries to keep dissimilar points far apart. 

t-SNE is a nonlinear dimensionality reduction technique developed by Laurens van der Maaten and Geoffrey Hinton~\cite{lvdm08}. $t$-SNE defines pairwise similarities $P$ and $Q$ on the input and output $\mathcal{X}$ and $\mathcal{Y}$, respectively, and it attempts to find $\mathcal{Y}$ so that these two measures of similarity match.

More specifically, the t-SNE algorithm computes input similarities as $p_{ij} = (p_{j | i} + p_{i | j})/2$, where 
\[
    p_{j | i} = \dfrac{\exp(-\Vert x_i - x_j \Vert^2 / 2\sigma_i^2)}{\sum_{k \neq i} \exp(-\Vert x_i - x_k \Vert^2 / 2\sigma_i^2)}.
\]
$\sigma_i$ are tuned so that the conditional distributions of $p_{j|i}$ have a user-specified behavior essentially controlling the number of nearest neighbors of $i$ supported on this distribution. van der Maaten and Hinton note that t-SNE is ``fairly robust to changes in the perplexity" \cite{lvdm08}, we will assume it to be fixed. We denote the matrix of pairwise similarities over $\mathcal{X}$ as $P \in \RR^{n \times n}$, where $(P)_{ij} = p_{ij}$.

The output similarities $q_{ij}$ are defined as
\[
    q_{ij} = \dfrac{(1 + \Vert y_i - y_j \Vert^2)^{-1}}{\sum_{k \neq \ell} (1 + \Vert y_k - y_\ell \Vert^2)^{-1}}.
\]
Letting $Q$ denote the matrix of similarities over $\mathcal{Y}$, the algorithm uses gradient descent to minimize the cost function over the embedding space:
\[
    C = \text{KL}(P || Q) = \sum_i \sum_j p_{ij} \log\dfrac{p_{ij}}{q_{ij}}
\]
where the gradient is
\[
    \dfrac{\partial C}{\partial y_i} = 4\sum_j (p_{ij} - q_{ij})(y_i - y_j)(1 + \Vert y_i - y_j \Vert^2)^{-1}.
\]

We can interpret the iterative optimization step of t-SNE (as well as that of many other dimensionality reduction methods) as an interaction between attractive and repulsive forces. Let us define $Z = \sum_{\ell=1}^n \sum_{k \neq \ell} (1 + \Vert y_k - y_\ell \Vert^2)^{-1}$. For the gradient descent step of t-SNE, we have
\begin{equation}
    \dfrac{1}{4}\dfrac{\partial C}{\partial y_i} = \underbrace{\sum_{j \neq i} p_{ij}q_{ij}Z(y_i - y_j)}_{\text{attraction}} - \underbrace{\sum_{j \neq i} q_{ij}^2Z(y_i - y_j)}_{\text{repulsion}}.
\label{eqn:attract_repulse}
\end{equation}
We refer to these kinds of methods as force-based methods, their precise form depends on the method used but most methods minimizing some notion of energy or similarity can be written in this form. For t-SNE, the expressions for these forces are found in Equation~\ref{eqn:attract_repulse}.

Kobak and Linderman~\cite{vsumap} found that t-SNE and UMAP produce similar results when used with the same initializations. Linderman and Steinerberger \cite{linderman} also noticed that t-SNE is asymptotically equivalent to a spectral clustering method. This suggests that all of these techniques should perhaps be considered as members of a family of force-based methods. A recent study by Bohm et al.~\cite{bohm} shows that embeddings generated via UMAP, ForceAtlas, and Laplacian eigenmaps can be empirically recovered by adjusting the balance of attractive and repulsive forces, suggesting that these methods lie along an attraction-repulsion spectrum.

\subsection{Spectral methods: a simple example}
\label{sec:spectral}
Let us analyze the a simple case of spectral embeddings. Suppose that we have constructed a graph $G = (V,E)$ from our data set, where $n = |V|$, vertices represent data points, and two vertices are connected by an edge if, for example, two data points are sufficiently close. We wish to identify functions $f: V \rightarrow \RR$ that can serve as coordinates on the data and we posit that such functions should vary by only a small amount over vertices that are connected by an edge. Such a framework leads to the natural optimization problem; find an $\hat f$ with
\begin{equation}
    \hat f = {\rm argmin}_{f:V \rightarrow \RR \atop {\|f\|_{2} = 1 \atop \sum_{v \in V} f(v) =0}} \sum_{(u,v) \in E} (f(u) - f(v))^2.
\label{eqn:energy_diff}
\end{equation}
Let us identify a function $f: V \rightarrow \RR$ with a vector and observe that
\[
    \sum_{(u,v) \in E} (f(u) - f(v))^2 = \left\langle f, (D-A) f \right\rangle,
\]
where $D \in \mathbb{R}^{n \times n}$ is the diagonal matrix $D_{ii} = \deg(v_i)$ and $A \in\mathbb{R}^{n \times n}$ is the adjacency matrix of the graph $G$. The minimizer of Equation~\ref{eqn:energy_diff} is the eigenvector of $D - A$ corresponding to the smallest nontrivial eigenvalue. The higher-order eigenvectors are suitable representations that are all orthogonal to each other (this is the intuition behind Laplacian eigenmaps).

Let us consider the solution to Equation~\ref{eqn:energy_diff} a little bit differently. Take such an eigenvector $f$ and consider the quadratic form 
\[
    \left\langle f, (D-A)f \right\rangle = \lambda \|f\|^2.
\]
We note that, as a function from $\mathbb{R}^n \rightarrow \mathbb{R}$, since $D-A$ is symmetric, the quadratic form has a simple gradient
\[
    \nabla  \left\langle x, (D-A) x \right\rangle = 2 (D-A)x.
\]    
Therefore, if we are trying locally to minimize the energy of our point configuration, we would like to move in the direction of the negative gradient and consider the new point $x_{\varepsilon} = x - \varepsilon 2 (D-A)x$. Taking the $i-$th entry and using the definition of $D$ and $A$, we see that
\[
     (x_{\varepsilon})_i = x_i - 2 \varepsilon  \left[(D-A) x\right]_i = x_i + 2 \varepsilon \sum_{(i,j) \in E} (x_j - x_i).
\]
This means that in classical Laplacian eigenmaps, each individual point $x_i$ has attractive forces that move it in direction of those points to which it is adjacent in the graph while it is held in place by both the orthogonality conditions and the $\ell^2-$normalization which ensures points are not clustered too close to the origin. When each position is the coordinate of an eigenvector, \textit{these attractive forces are simply proportional to a point's position in space and do not carry any additional information} as $ \left[(D-A) f \right]_i  = \lambda_i f_i$. We note that this is not at all the case for embedding methods such as t-SNE, in which the attractive forces on each point (see Equation~\ref{eqn:attract_repulse}) are carry additional information that we wish to exploit.

\section{Vector Flow Procedure} \label{sec:vector_flow}
We summarize the vector flow procedure in Algorithm \ref{alg:vector_flow}. Given a dataset $\mathcal{X} \subseteq \RR^s$, we first run t-SNE to equilibrium to generate an embedding $\mathcal{Y}_0$. We then perform spatial interpolation on $\mathcal{Y}_0$ and $\mathcal{\tilde F}$, a modification of the attraction forces at equilibrium, to generate a fixed vector field $\mathcal{V}$. Next, we flow the embedding along $\mathcal{V}$ for a user-specified number of iterations $T$.

We note that the vector field we use is not \emph{exactly} the attractive forces at equilibrium but, rather, a modification of the vector field of attractive forces at each point that provides a measure of the correlation between neighbors of a point $x_i$ in the input space and the neighbors of $y_j$ in the embedding space. The original attractive forces do not produce useful flows (Fig. \ref{fig:vector_flow:orig_flow} in the Appendix). By Clairaut's theorem, the modified attractive forces, unlike the tSNE dynamical system, are \emph{not} the gradient of an energy functional. They do, however, give rise to a nonlinear dynamical system in two dimensions on the embedded points the features of which are subject of ongoing work (e.g., mean-field approximation, interaction with Gaussian interpolation, etc.).

\begin{algorithm}[!ht]
\SetKwInput{Input}{Input}\SetKwInput{Output}{Output}
\DontPrintSemicolon 
  \Input{$\mathcal{X} \subseteq \RR^s$, $T$}
  \Output{$\mathcal{Y}_{flowed} \subseteq \RR^2$}
  Run t-SNE on $\mathcal{X} = \{x_1, \ldots, x_n\}$ to equilibrium to produce embedding $\mathcal{Y}_0 = \{y_1, \ldots, y_n\} \subseteq \RR^2$. \\
  Calculate $\mathcal{\tilde F} = \{f_1, \ldots, f_n\}$, where $f_i = \sum_{j \neq i} Z (y_i - y_j) \sum_{k \neq i} p_{ik} q_{kj} $ is a smoothed attraction force at equilibrium for each $y_i \in \mathcal{Y}_0$, \\
  $\mathcal{Y}_{flowed} \leftarrow \mathcal{Y}_0$ \\
  \For{$t \in \{1, \ldots, T\}$}{
    \For{$y \in \mathcal{Y}_{flowed}$}{
        $\text{Force}(y) \leftarrow$ SpatialInterpolation($y$, $\mathcal{Y}_0$, $\mathcal{\tilde F}$) \\
        $y \leftarrow y + \text{Force}(y)$
    }
  }
\caption{Vector flow procedure}
\label{alg:vector_flow}
\end{algorithm}

\begin{figure}[h!]
\centering
\includegraphics[width=8cm]{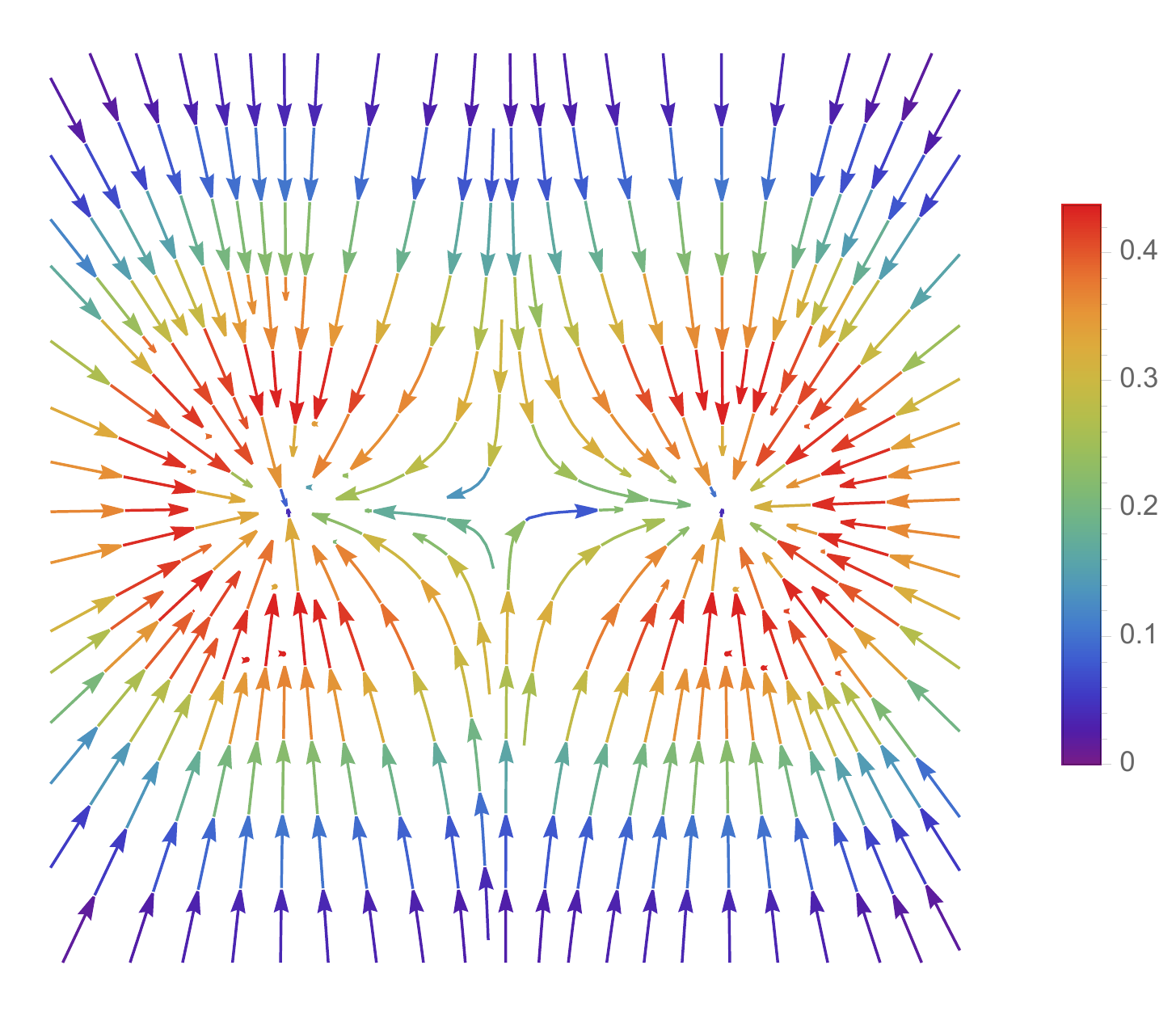}
\vspace{-5pt}
\caption{The resulting vector field $\mathcal{V}$, colored by magnitude, of two embedded Gaussian clusterings. $\mathcal{V}$ contains sinks around the cluster centers.}
\label{fig:vector_flow:gauss:vectors_c}
\end{figure}

In Fig.~\ref{fig:vector_flow:gauss:vectors_c}, we illustrate the vector field corresponding to the embedding of two Gaussian clusters. For more details about this particular embedding, see Section~\ref{sec:experiments}. The figure shows the direction and magnitude of the field, along which we flow the embedded points. Notice that there are sinks in the vector field near the centers of the clusters. There are also some asymmetries in the vector field (despite originating from a symmetric distribution) and some abrupt changes in magnitude and direction in the microstructure of the field. 

We interpolate the embedding forces using a Gaussian-weighted $k$-nearest neighbors average in all experiments. This calculation is described in Algorithm \ref{alg:interpolation}. The procedure depends on two hyperparameters: the number of neighbors $k$ and the Gaussian bandwidth $\sigma$. In order to ensure that the kernel assigns nontrivial weight over a local neighborhood of each point, we set $\sigma$ as the mean $5$-NN distance across all points in $\mathcal{Y}_0$. We then set $k$ to the minimum $\tilde{k}$ such that the mean $\tilde{k}$-NN distance over $\mathcal{Y}_0$ exceeds $2\sigma$. Investigating the effects of a more careful selection of these parameters is a worthwhile direction for future work.

\begin{algorithm}[!ht]
\SetKwInput{Input}{Input}\SetKwInput{Output}{Output}\SetKwInOut{Params}{Params}
\DontPrintSemicolon
  \Input{$y$, $\mathcal{Y}_0$, $\mathcal{F} \subseteq \RR^2$}
  \Output{$\text{Force}(y) \in \RR^2$}
  \Params{$k$, $\sigma$}
  $\mathcal{Y}_0 = \{y_1^{(0)}, \ldots, y_n^{(0)}\}$ is the embedding. $\mathcal{F} = \{f_1, \ldots, f_n\}$ are the corresponding forces. \\
  Find $\mathcal{N}(y)$ the $k$ nearest neighbors of $y$ in $\mathcal{Y}_0$. \\
  \For{$y_i^{(0)} \in \mathcal{Y}_{0}$}{
    Calculate the weight
    $$
    w(y, y_i^{(0)}) = \exp\left(-\tfrac{1}{2\sigma^2}\Vert y - y_i^{(0)} \Vert^2\right)
    $$
  }
  Return the weighted average of the embedding forces:
  $$
  \text{Force}(y) \leftarrow \tfrac{\sum_{y_i^{(0)} \in \mathcal{N}(y)} f_i \cdot w(y, y_i^{(0)})}{\sum_{y_i^{(0)} \in \mathcal{N}(y)} w(y, y_i^{(0)})}
  $$
\caption{Gaussian-Weighted Interpolation}
\label{alg:interpolation}
\end{algorithm}

We used a Gaussian weighted $k$-NN average because it is simple to implement, runs relatively quickly~\cite{fggt}, and provides a reasonable weight scaling over the neighborhood of $y$. We note that there are a number of alternative interpolation methods, including an inverse distance method, a generic RBF kernel (as opposed to a Gaussian kernel), and barycentric mesh interpolation. The last interpolation scheme is mesh-based, while the others are mesh-free (and might be considerably faster)~\cite{solomon}. 
\section{Experiments}
\label{sec:experiments}

We performed four classes of experiments to test the empirical performance of our algorithm. We begin with a toy example, two Gaussians, a simple example where our method works as expected. We see two well-separated clusters flow to two sinks, with decreasing cluster separation eventually resulting in failure. Next, we investigate two MNIST datasets and showcase several examples in MNIST and Fashion-MNIST where the vector field flows reveal interesting substructures and subfeatures in the data. These experiments indicate that flowing along the attractive forces vector field may have useful application for real data. In order to determine where this substructure comes from and if it is indicative of real structure in the data, we perform an experiment with a single Gaussian cluster, an example where there should be no substructure. We see evidence of artifacts as a result of the microstructure of the vector field. Finally, we show that we can average out the microstructure of the vector field to provide consistent view points and consistent features. In other words, we show that running t-SNE several times on the same data set (using random initializations) can provide a consistent view of the data, a property that is sorely missing in the current applications of t-SNE.

Unless specified otherwise, all of our experiment ran $t$-SNE with a PCA initialization. This is the default setting for the \href{https://opentsne.readthedocs.io/en/latest/index.html}{openTSNE} Python package \cite{openTSNE}. Please see the Appendix for more information on implementation details and parameters for reproducibility.

\subsection{Toy Gaussians}
\begin{figure}[h!]
\centering
\includegraphics[width=4cm]{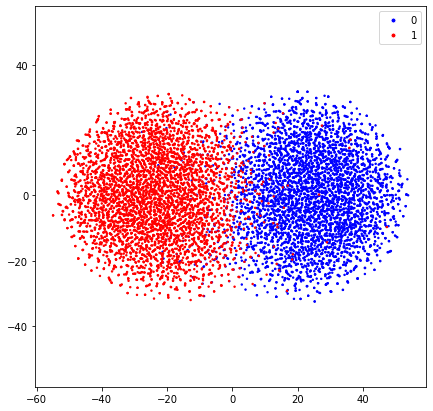}
\includegraphics[width=4cm]{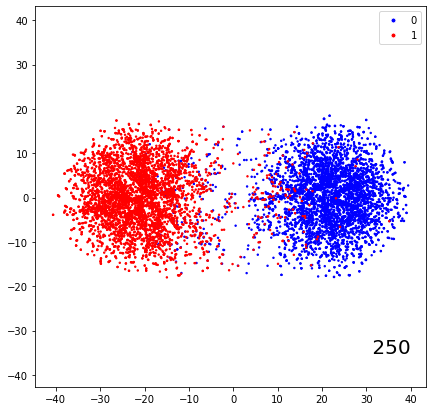}
\includegraphics[width=4cm]{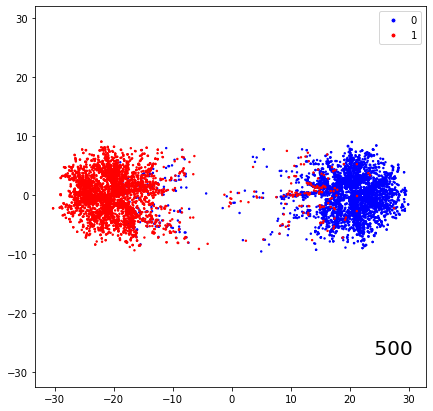}
\vspace{-5pt}
\includegraphics[width=4cm]{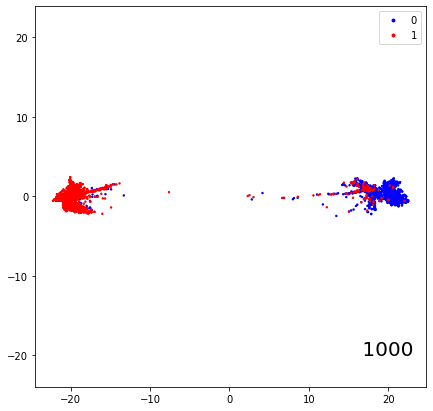}
\vspace{-5pt}
\caption{Embedding with ground truth coloring coloring (top left) flowed along the vector field in Fig.~\ref{fig:vector_flow:gauss:vectors_c} up to $1000$ iterations. Points flow towards the force sinks in Fig.~\ref{fig:vector_flow:gauss:vectors_c}, makingthe clusters more clearly distinguishable.}
\label{fig:vector_flow:gauss}
\end{figure}

We first tested the vector flow procedure on a toy dataset sampled from two high-dimensional Gaussian clusters. The vector field $\mathcal{V}$ generated from the embedding has a sink near the centers of each cluster (Fig. ~\ref{fig:vector_flow:gauss:vectors_c}). We observe that the embedded clusters converge to these sinks when flowed along $\mathcal{V}$ (Fig. ~\ref{fig:vector_flow:gauss}). This demonstrates that attraction flows improve the visibility of ground truth clusters for this simple example. 


To quantitatively evaluate the effects of flowing, we compared the performance of $k$-means clustering, with $k$ fixed at $2$, on the flowed embedding versus the original embedding. We sampled data from two Gaussian clusters in $\RR^{20}$ with unit covariance using $2000$ points per cluster. We flowed the vector field for 1000 iterations and then ran $k$-means. 
As we increased the intra-cluster distance of the ground truth distributions, we observed that $k$-means is roughly equally effective for cluster identification on the original and flowed embeddings. For both embeddings, $k$-means is less effective when the intra-cluster distance is small and is highly effective when the clusters are well-separated. This suggests that attraction flows do not worsen cluster quality on this type of example. We also observed that flows on embeddings generated from distributions with higher intra-cluster distance converge in fewer iterations.

\subsection{Feature identification and cluster substructure}

The unlabeled t-SNE embeddings for datasets with low ground-truth separation distance indicate the presence of only a single cluster, yet the flows appear to reveal substructures related to true two-cluster distribution. To show that attraction flows can be applied to more complex datasets to elucidate information not present in ordinary t-SNE output, we apply our algorithms to MNIST \cite{MNIST} and Fashion-MNIST datasets \cite{fashion}.  

\subsubsection{MNIST digits}
\begin{figure}[h!]
\centering
\includegraphics[width=2.75cm]{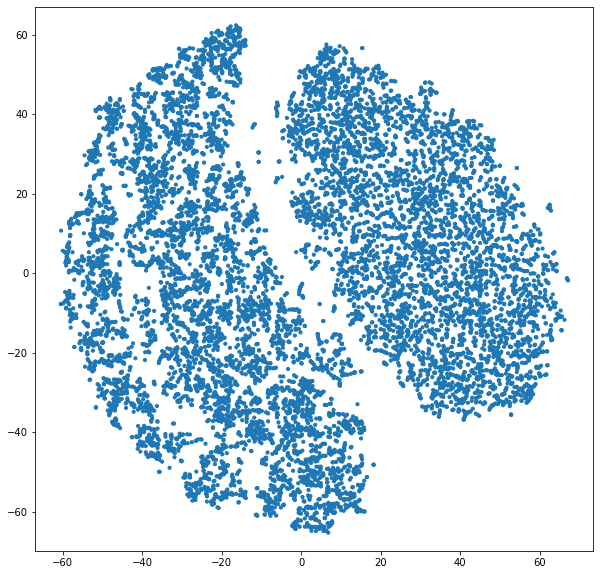}
\includegraphics[width=2.75cm]{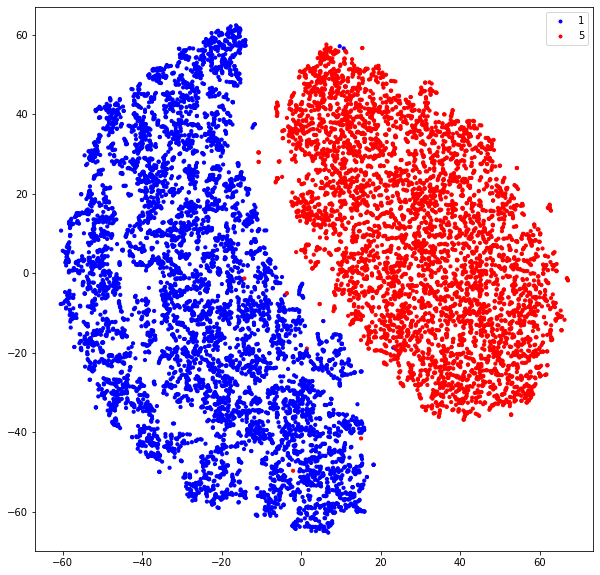}
\includegraphics[width=2.75cm]{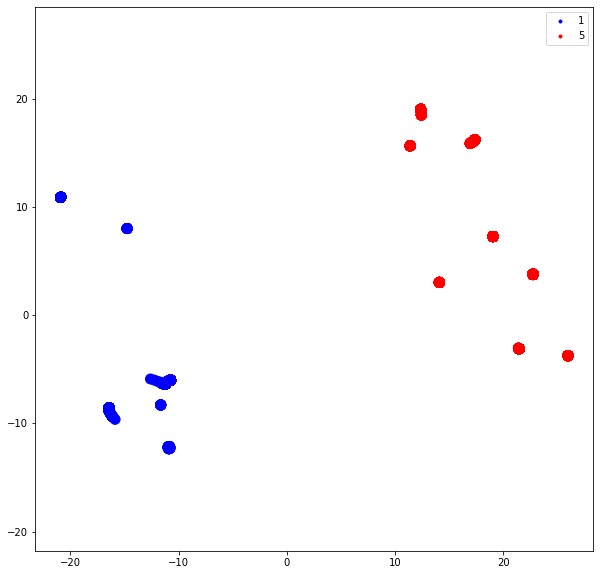}
\vspace{-5pt}
\caption{Embedding of digits $1$ and $5$ from MNIST handwritten digits dataset. Unlabeled embedding (left), ground truth labels (center), embedding after $40000$ flow iterations (right). Digits $1$ are colored blue, digits $5$ are colored red.}
\label{fig:vector_flow:digits:blind}
\end{figure}

We examined the attraction flows on the MNIST handwritten digits dataset \cite{MNIST}, focusing specifically\footnote{We performed other experiments, not shown, on difficult to disambiguate digits (e.g., 4 and 9) and saw similar results. For the sake of exposition, we stick with 1 and 5 as their features are easy to describe.} on embeddings of the digits $1$ and $5$ (Fig. \ref{fig:vector_flow:digits:blind}). As we can see, the initial t-SNE embedding appears to show two large clusters--however, it is entirely unclear whether the rich patterns within each cluster correspond to meaningful features as well. After flowing points along the vector field for $40000$ iterations, the embedding converges to a set of sinks that unambiguously partition the samples. Despite there being only two classes present in the data, the vector field has over ten sinks.

\begin{figure}[h!]
\centering
\includegraphics[width=4cm]{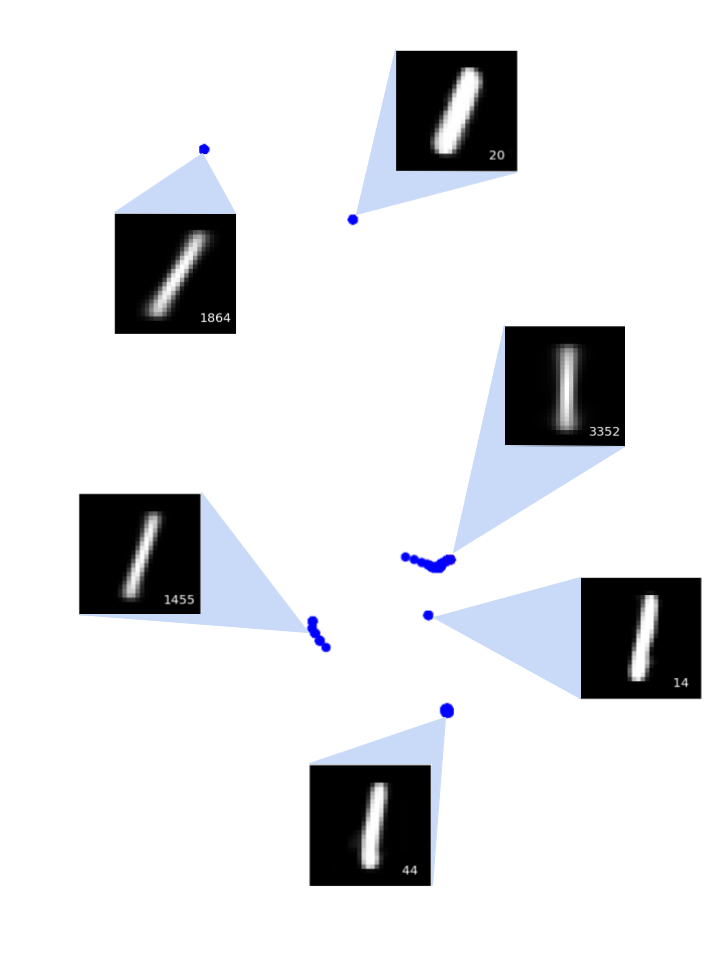}
\includegraphics[width=4cm]{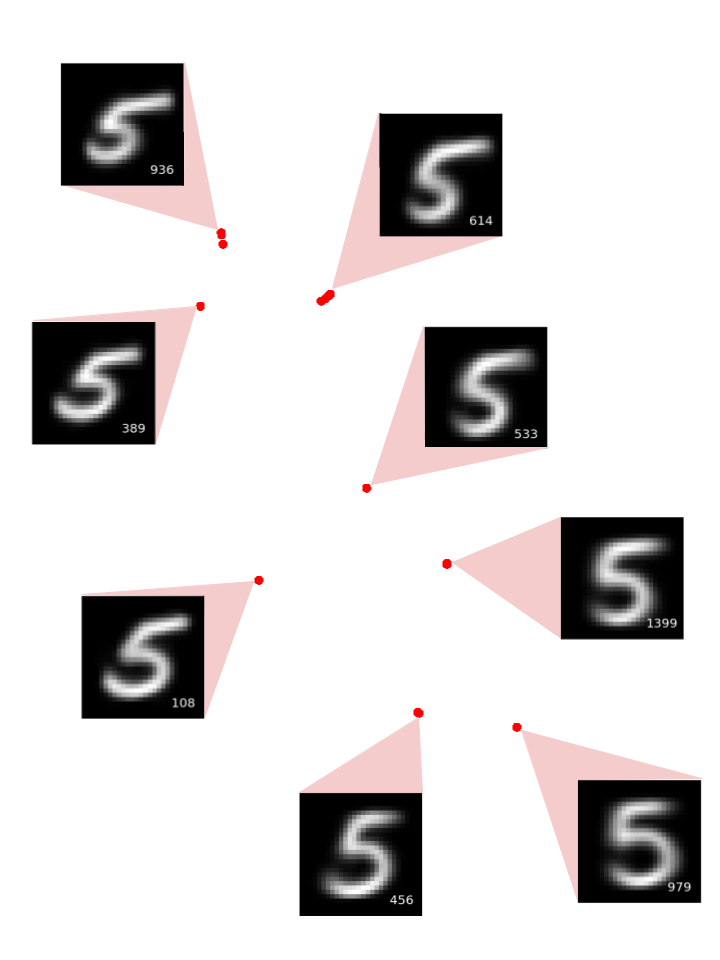}
\vspace{-5pt}
\caption{Closeup of flowed digits $1$ (left) and 5 (right) from Fig~\ref{fig:vector_flow:digits:blind} labeled with cluster means. Number of images per cluster is shown in white. The clusters correspond to digits $1$ and digits $5$ written in various styles.}
\label{fig:vector_flow:digits:40k_15}
\end{figure}

To determine whether these sinks are indicative of additional features, we examine the average image embedded in each cluster. This is shown in Fig.~\ref{fig:vector_flow:digits:40k_15}. We observe that each sink consists of digits with distinct features. For example, the sinks corresponding to the digits $1$, shown in blue on the right, vary in thickness and slant across clusters. The digits $5$, shown in red on the left, vary in the size and slant of their bottom hooks and top dashes. Thus, the clusters indicate different handwriting styles.

\begin{figure}[h!]
\centering
\includegraphics[width=6cm]{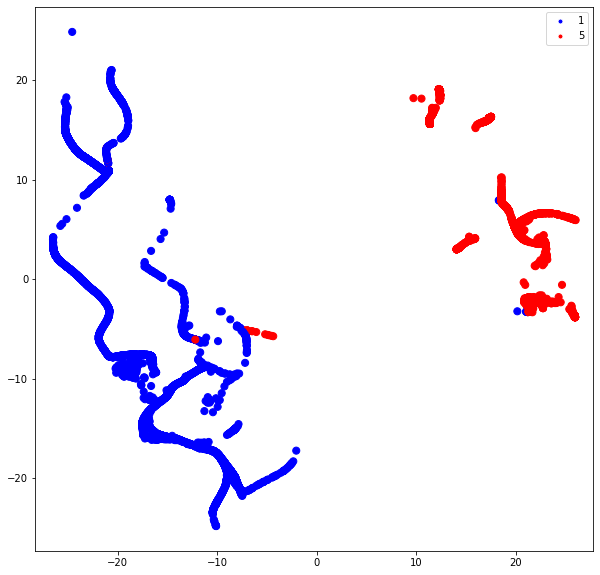}
\vspace{-5pt}
\caption{Closeup of flowed embedding from Fig~\ref{fig:vector_flow:digits:blind} after $5000$ iterations. Blue is digit $1$, red is digit $5$. }
\label{fig:vector_flow:digits:5k}
\end{figure}

We also observed that intermediate flows reveal interesting information about the underlying structure of the data. In Fig.~\ref{fig:vector_flow:digits:5k}, we show the embedding after after $5000$ iterations, colored by ground truth. The tendril-like trajectories in the embedding organize the digits along a continuum of subfeatures. 

\begin{figure}[h]
\centering
\includegraphics[width=8cm]{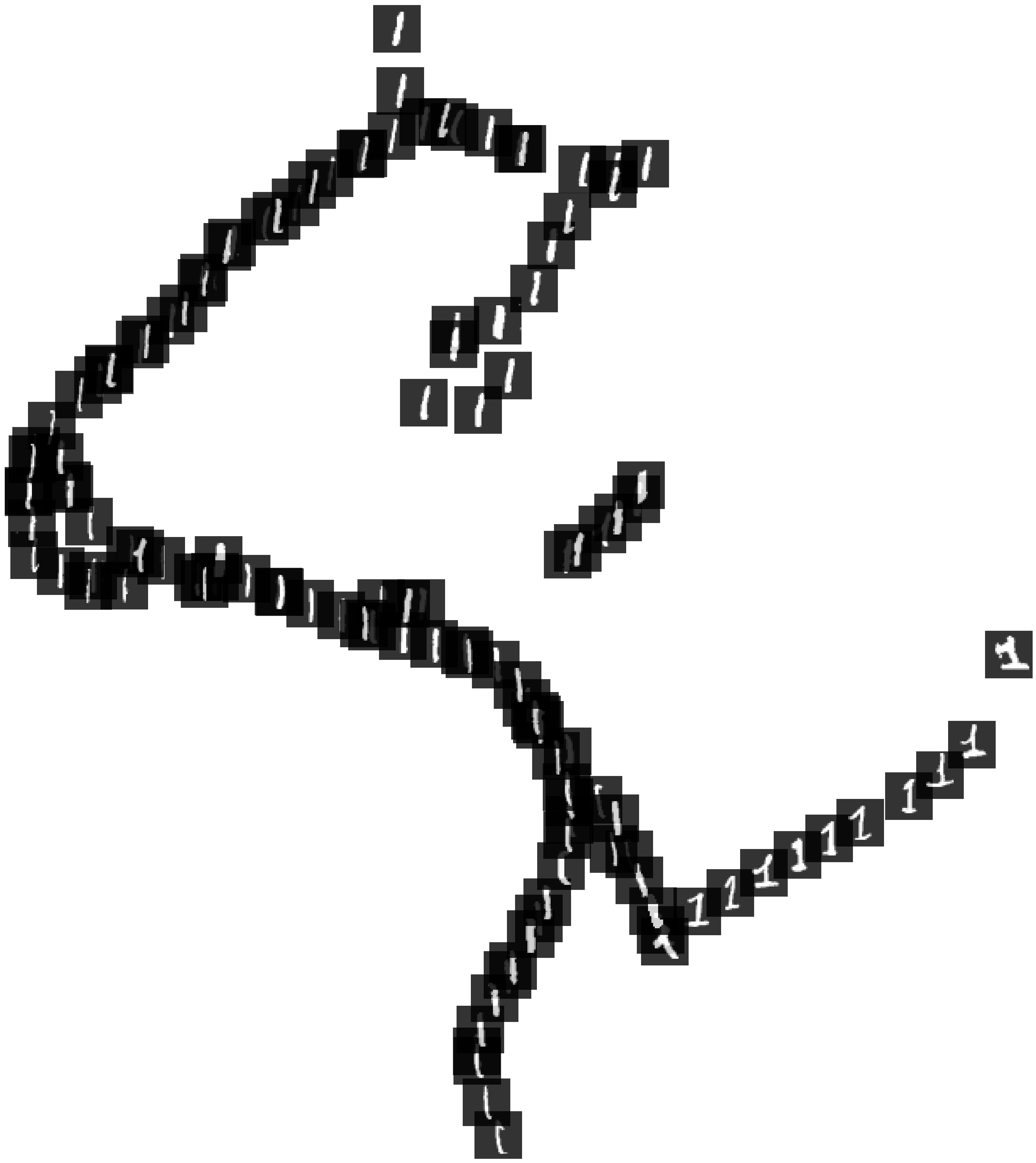}
\vspace{-5pt}
\caption{Segment of blue flow lines from Fig~\ref{fig:vector_flow:digits:5k}, corresponding to the digit 1.}
\label{fig:vector_flow:digits:spectrum_1}
\end{figure}

\begin{figure}[h]
\centering
\includegraphics[width=9cm]{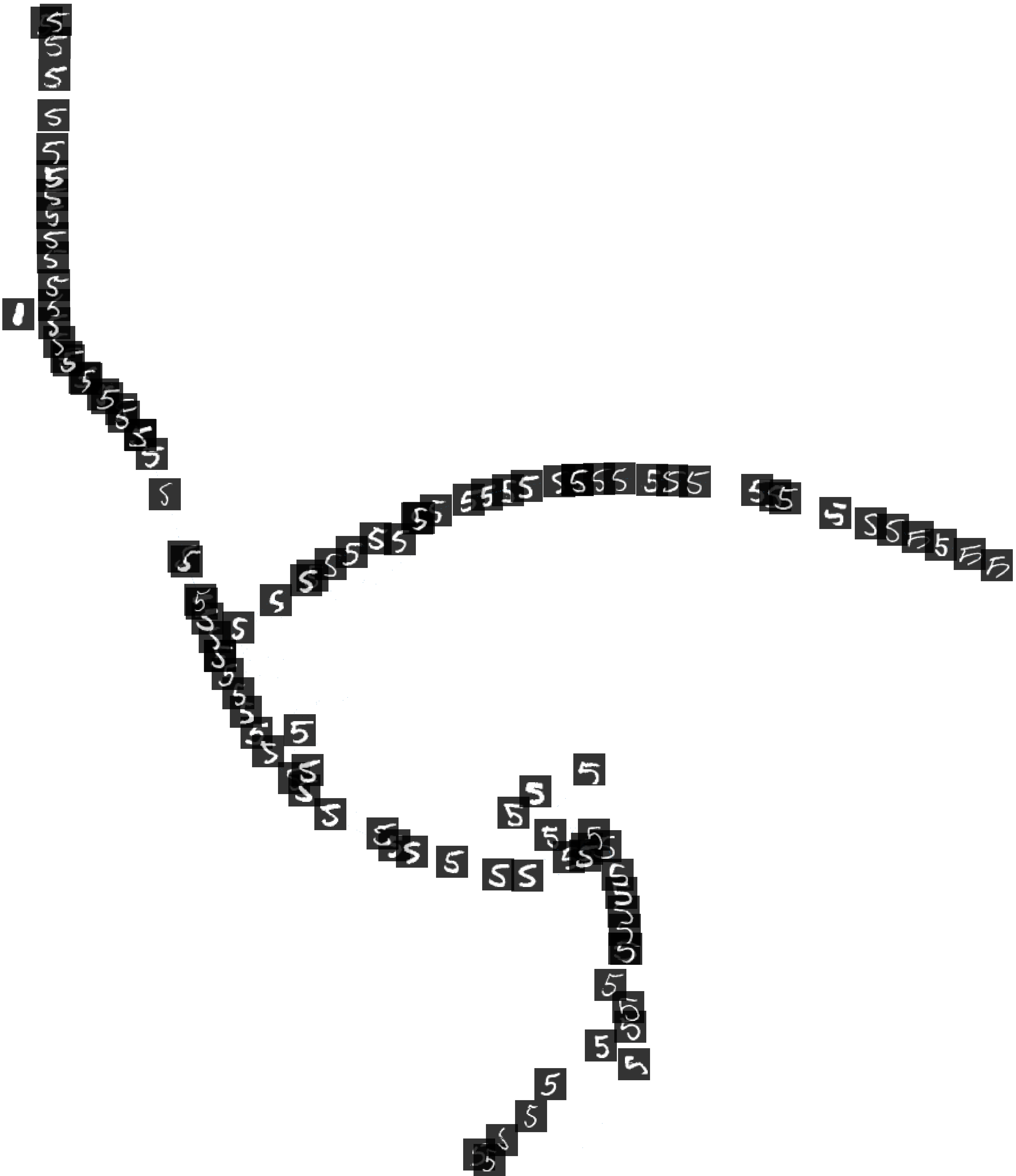}
\vspace{-5pt}
\caption{Segment of red flow lines from Fig~\ref{fig:vector_flow:digits:5k}, corresponding to the digit 5.}
\label{fig:vector_flow:digits:spectrum_5}
\end{figure}

\begin{figure}
\centering
\includegraphics[width=3cm]{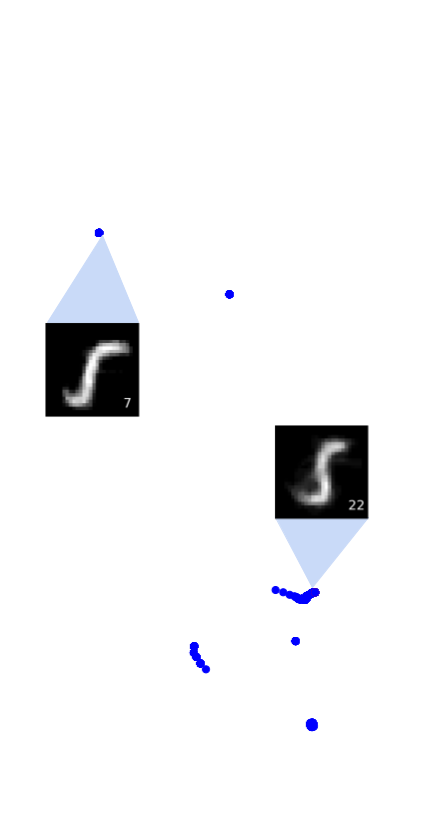}
\includegraphics[width=3.5cm]{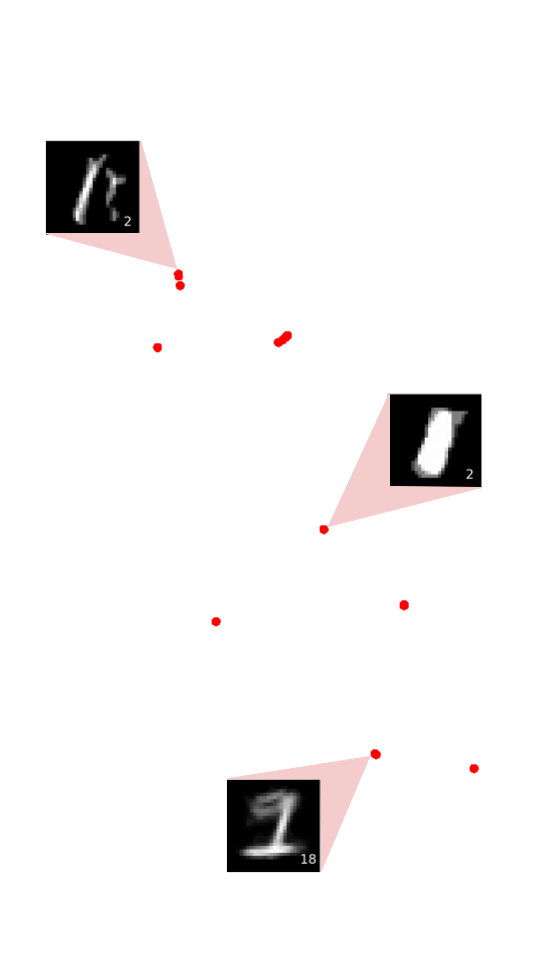}
\vspace{-5pt}
\caption{Misclassified 1's and 5's. Clusters in the flowed embedding above are labeled with their mean misclassified digit, if present. We found that $22$ out of ~$6700$ digit $1$ samples and $22$ out of ~$5400$ digit $5$ samples were misclassified.}
\label{fig:vector_flow:digits:misclassified}
\end{figure}

We illustrate this finding with the examples in Figs.~\ref{fig:vector_flow:digits:spectrum_1} and~\ref{fig:vector_flow:digits:spectrum_5}. Following the flow line in Fig.~\ref{fig:vector_flow:digits:spectrum_1}, the digits $1$ are clearly organized into a spectrum based on stroke weight and angle. In the top right, digits are written with a heavy stroke and slight rightward slant. Following the flow line downwards, the digits become thinner and transition to a neutral slant. At the bottom, the line branches into two forks. The left fork contains digits with a neutral to slight leftward slant. At the end we see that the digits have a convex curve at the bottom. The right fork contains digits that become thicker and have with a much more severe leftward slant. The right fork is also joined by a group of digits $1$ with serifs. At the bottom we see a thick stroke digit with a serif. There are also messy $5$'s (buried under other images) like those shown in ~\ref{fig:vector_flow:digits:misclassified}. This increased whitespace at the top and bottom of the digits transitions the line to a group of digits with serifs and bases. 

As in Fig~\ref{fig:vector_flow:digits:spectrum_1}, the flow lines in Fig~\ref{fig:vector_flow:digits:spectrum_5} appear to organize the digits $5$ on a spectrum. These lines ultimately converge to two sinks: one at the intersection of the horizontal and long vertical branches and one at the corner of the vertical branch in the lower right of the image. As we follow the horizontal branch from right to left, the stroke weight of the digits generally becomes thicker. We observe a similar transition as we follow the vertical branch from the top to the the lower sink. As we follow the vertical branch from the bottom to the lower sink, we observe that digits transition from being narrow with a thin stroke to being wide with a thick stroke.

The majority of digits $1$ are embedded in the left cluster, while the majority of digits $5$ are embedded in the right cluster, as shown in Fig. \ref{fig:vector_flow:digits:blind}. However, a handful of points are mixed up between these two clusters, even after flowing. These are shown in Fig.~\ref{fig:vector_flow:digits:misclassified}, and we observe that they generally correspond to digits which are written messily or unconventionally. For example, the leftmost group of misclassified $5$'s have very flat hooks and dashes, making them look similar to $1$'s even with human eyes. Similarly, the bottom group of misclassified $1$'s have large serifs and bases that could plausibly be mistaken for the dash and the bottom part of a hook for a $5$.

We also tried flowing an embedding of the digits $5$ alone. This produced only $4$ subclusters even though we observed around $8$ for the embedding of this digit with $1$. It is not clear why the presence of a second class drastically distorts the number of subclusters found. We hope to address this in future work.


\subsubsection{Fashion MNIST}

\begin{figure}[h!]
\centering
\includegraphics[width=2.75cm]{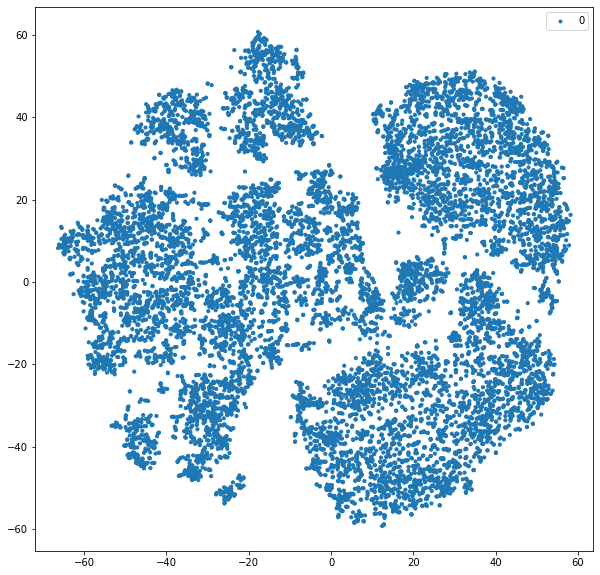}
\includegraphics[width=2.75cm]{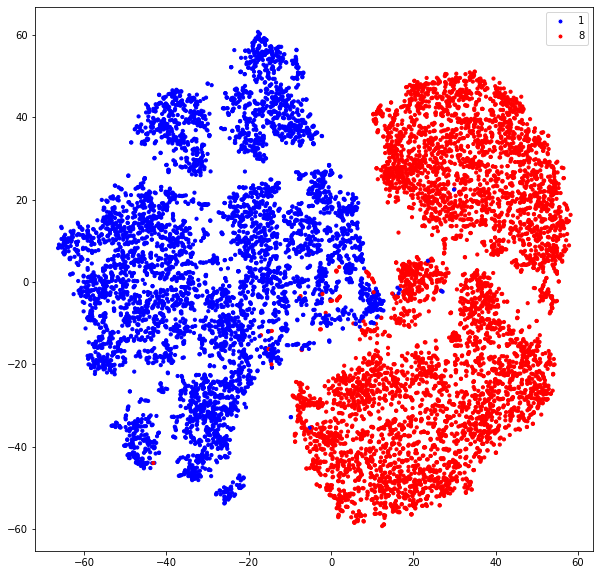}
\includegraphics[width=2.75cm]{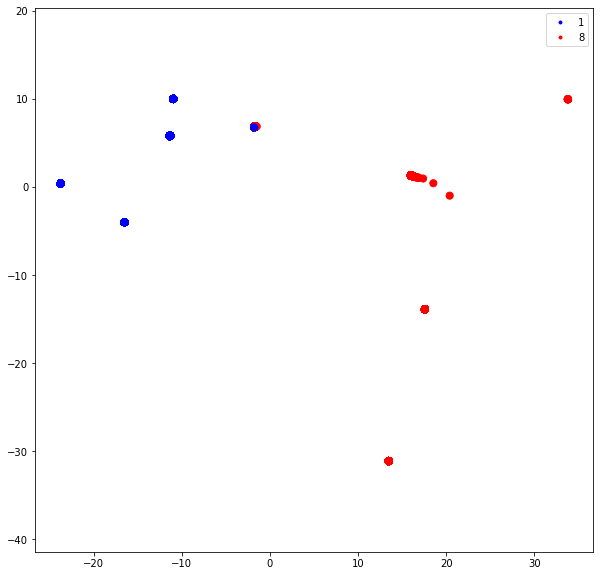}
\vspace{-5pt}
\caption{Embedding of pants and purses from Fashion-MNIST dataset. Unlabeled embedding (left), ground truth labels (center), embedding after $20000$ flow iterations (right). Pants are colored blue, purses are colored red.}
\label{fig:vector_flow:pants:blind}
\end{figure}

We performed a similar analysis of Fashion MNIST \cite{fashion}, which consists of images of clothing. We first examined the joint embedding of images of pants and purses (Fig \ref{fig:vector_flow:pants:blind}).

\begin{figure}[h!]
\centering
\includegraphics[width=6cm]{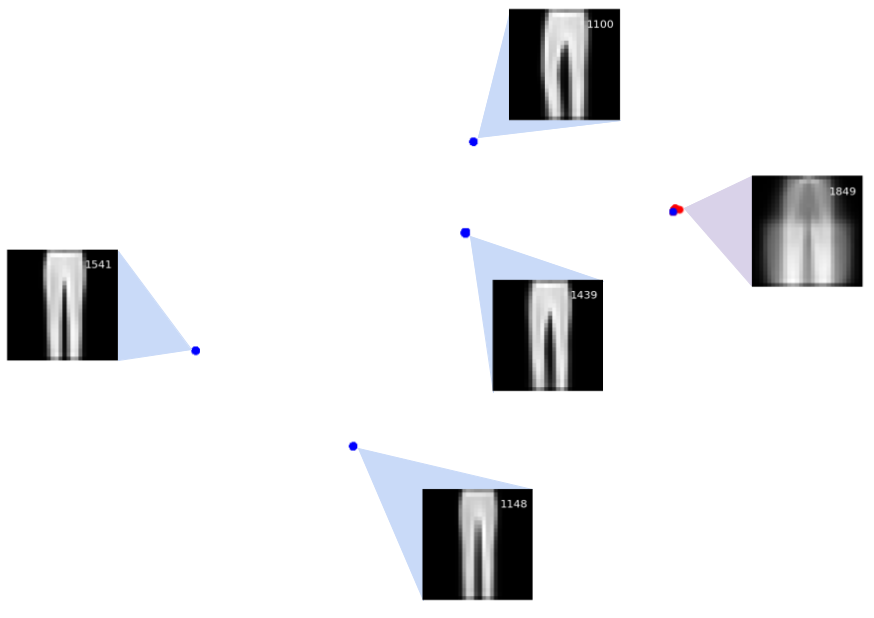}\\
\includegraphics[height=5cm]{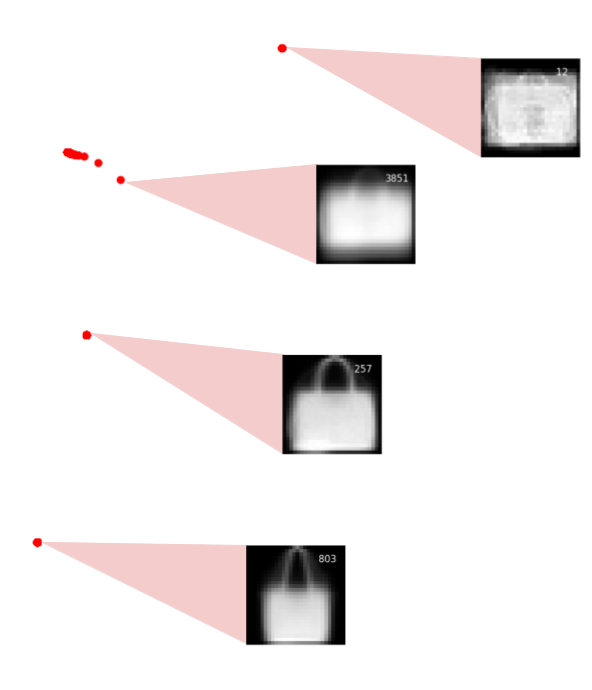}
\vspace{-5pt}
\caption{Closeup of flowed pants samples (left) and purse samples (right) from Fig~\ref{fig:vector_flow:pants:blind} labeled with cluster means. Number of images per cluster is shown in white. The clusters correspond to pants of different width and leg orientation and purses of different shapes. The cluster in the middle is mixed up with purses; see the discussion of Fig~\ref{fig:vector_flow:pants:transition}.}
\label{fig:vector_flow:pants:20k_18}
\end{figure}

As seen in Fig \ref{fig:vector_flow:pants:blind}, the clusters are well separated, with the majority of pants images embedded in the left cluster and the majority of purse images embedded in the right. We flowed the embedding for $20000$ iterations so that the images converged to sinks, and we noticed that one sink in the middle attracted a mixture of samples from both classes. To examine this further, we again plotted the mean image in each sink. This is shown in Fig~\ref{fig:vector_flow:pants:20k_18}.

As in the MNIST digits example, the limiting clusters indicate subfeatures of the data. For the pants, shown in blue on the right, the clusters indicate different widths and positions of the pant legs. For the purses, shown in red on the right, the clusters indicate different bag shapes and handle lengths. The cluster that combined samples from both classes, shown in purple on top, is clearly mixed.

\begin{figure}[h!]
\centering
\includegraphics[width=8cm]{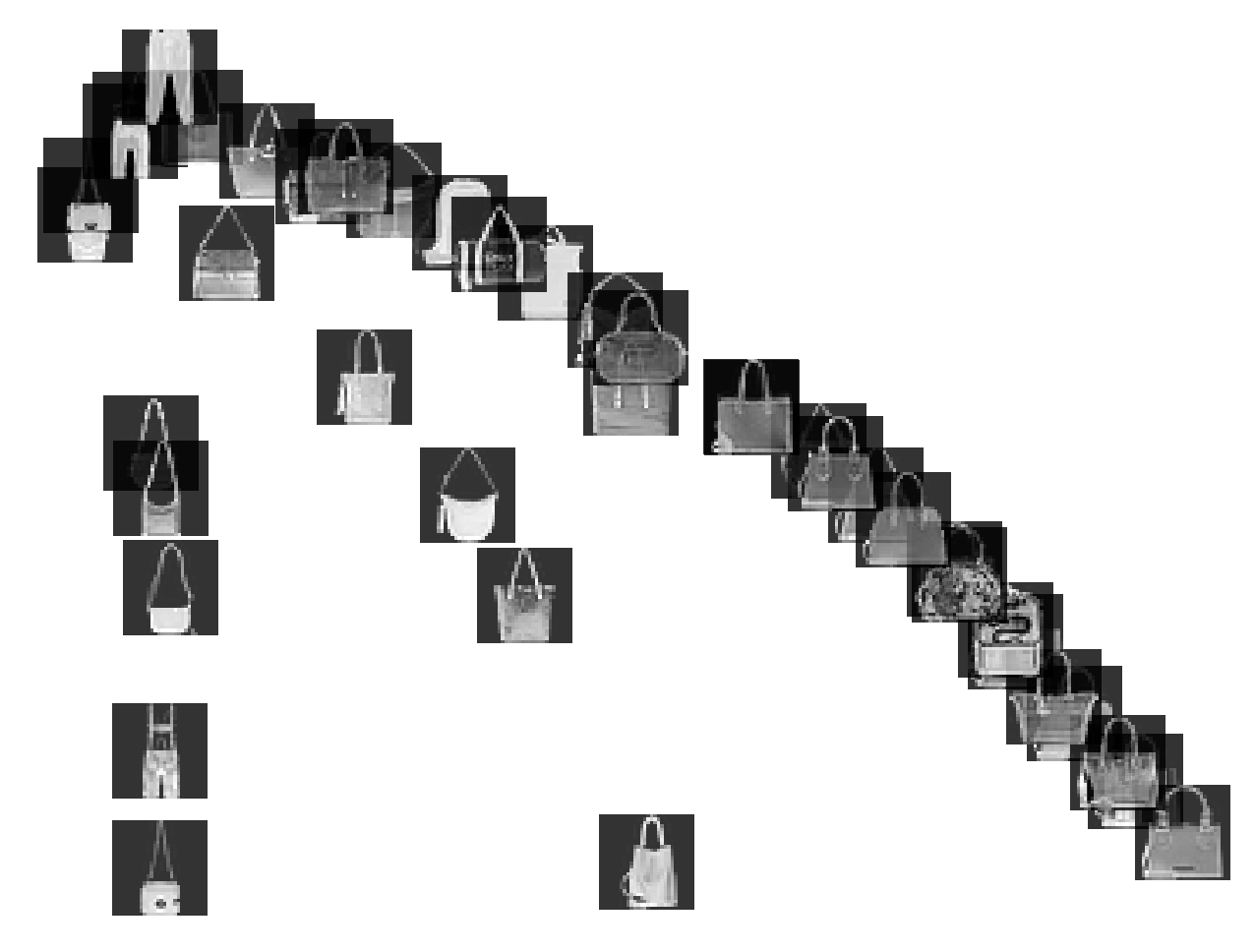}
\vspace{-5pt}
\caption{Why pants and purses were clustered together in Fig~\ref{fig:vector_flow:pants:20k_18}. The pants inseam has roughly the same shape as a purse handle; the flow lines above appear to organize the data into a spectrum based on this subfeature. The right streamline consists mostly of bulky handbags with a short handle. In the top left corner, these handles transition into pants. Following the left streamline down the plot, the pants transition into the long strap of a crossbody purse.}
\label{fig:vector_flow:pants:transition}
\end{figure}

We examine this mixed cluster in greater detail on the intermediate flow after $10000$ iterations. This is shown in Fig.~\ref{fig:vector_flow:pants:10k} of the Appendix. We show a closeup of the region with the mixed cluster in Fig~\ref{fig:vector_flow:pants:transition}. Quite interestingly, the combination of these two classes highlights an alternative subfeature of the data--the $\wedge$-shaped whitespace of the purse handle! Purses with prominent handles were mixed with pants because the handles look similar to the inseam of the pants. We also observed that the large purses in the cluster are relatively dark, whereas purses in other parts of the embedding tend to be light (Fig ~\ref{fig:vector_flow:pants:20k_18}, \ref{fig:vector_flow:pants:purses}). Since pants images contain substantial black space, this led the dark, bulky handbags to be interpreted as similar. We observed that the bulky purses, which comprise the majority of the purses in this cluster, were not mixed with pants in the joint embedding of pant, purses, and dresses.

Examining the streamlines for the light colored purses also revealed that the streamlines organize samples by subfeatures. See Fig~\ref{fig:vector_flow:pants:purses} in the Appendix. We performed analyses on other examples that provided similar evidence that flowing reveals information about data substructure, e.g. shirts and sweaters, sandals sneakers and boots, etc. However, these are omitted for space.

\subsection{Local microstructure}
In the previous section, we observed that the arrangement of images along flow trajectories during the process of flowing seems related to the subfeature structure of the embedding. We observe that we can obtain similar path-like trajectories in the two Gaussians dataset. 

\begin{figure}[h!]
\centering
\includegraphics[width=4cm]{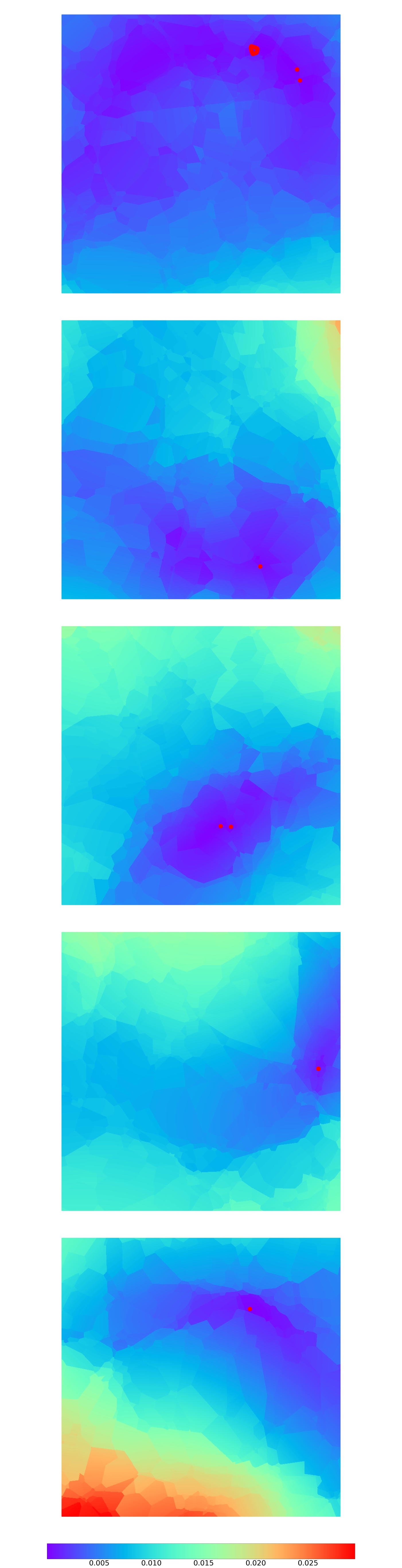}
\includegraphics[width=4cm]{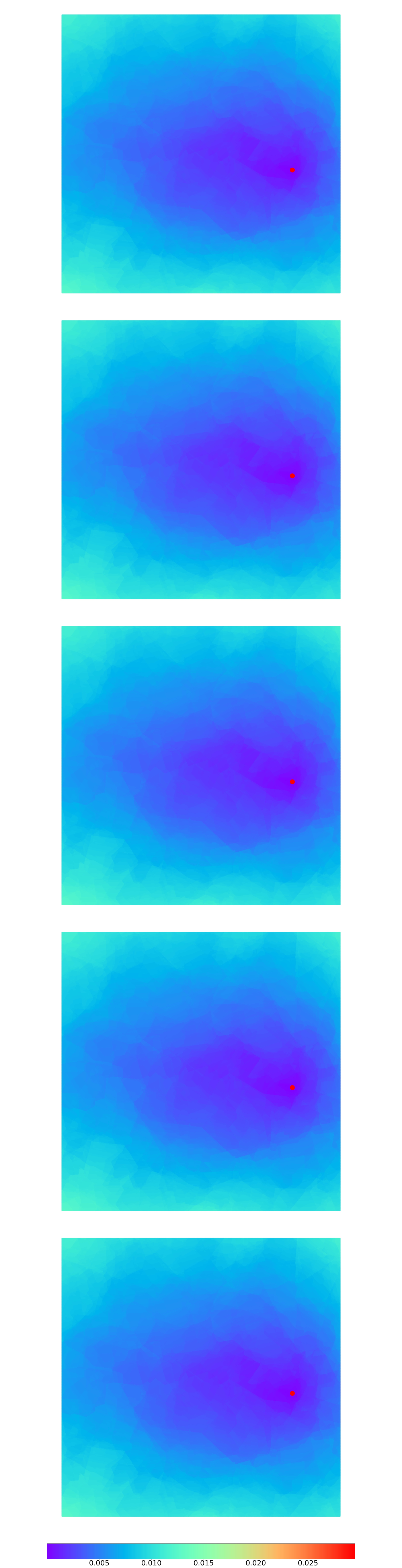}
\vspace{-5pt}
\caption{Embeddings of single Gaussian cluster generated from $5$ trials of $t$-SNE with random initialization and then flowed for $20000$ iterations (left). $2000$ samples per trial. Averaging the interpolated vector fields for the five trials produces a mean vector field with a single sink towards which all trials flowed (right). All plots show the subset $[-2,2]^2 \subseteq \RR^2$ and use the same colormap for vector magnitudes. Flowed embeddings are shown in red.}
\label{fig:vector_flow:random_init:gauss}
\end{figure}

We also observed that the addition or removal of a class of images affects the number and type of subclusters found. It is well known that t-SNE has a tendency to find patterns in random noise \cite{wattenberg} and that the method of initialization can drastically affect the final embedding \cite{vsumap}. In order to investigate whether the subclusters generated by attraction flows represent real substructure or artifact (either of the flow algorithm or t-SNE itself), we embedded a simple fixed dataset using t-SNE with random initialization and compared the flowed subclusters across several trials. We began by examining a single Gaussian cluster. We ran $5$ trials and flowed for $20000$ iterations in each. The flowed embeddings and their corresponding vector fields are shown in the left column of Fig.~\ref{fig:vector_flow:random_init:gauss}. The vector field is colored by force magnitude, and the embedding is plotted in red. We observed that the number of limiting sinks observed varies depending on the initialization. For instance, the first trial resulted in a limit cycle and two sinks, the third trial resulted in two sinks, but the remaining three trials all converged to a single sink. This is related to the fact that the lay of the vector field also varies substantially from run to run. 

Since the underlying distribution is radially symmetric with no substructure, we found that averaging the interpolated vector fields for the five trials produces a mean vector field with a single sink 
Flowing the randomly sampled embeddings on the mean vector field resulted in a single sink for all trials. This is shown on the right in Fig.~\ref{fig:vector_flow:random_init:gauss}. 

\begin{figure}[h!]
\centering
\includegraphics[width=4cm]{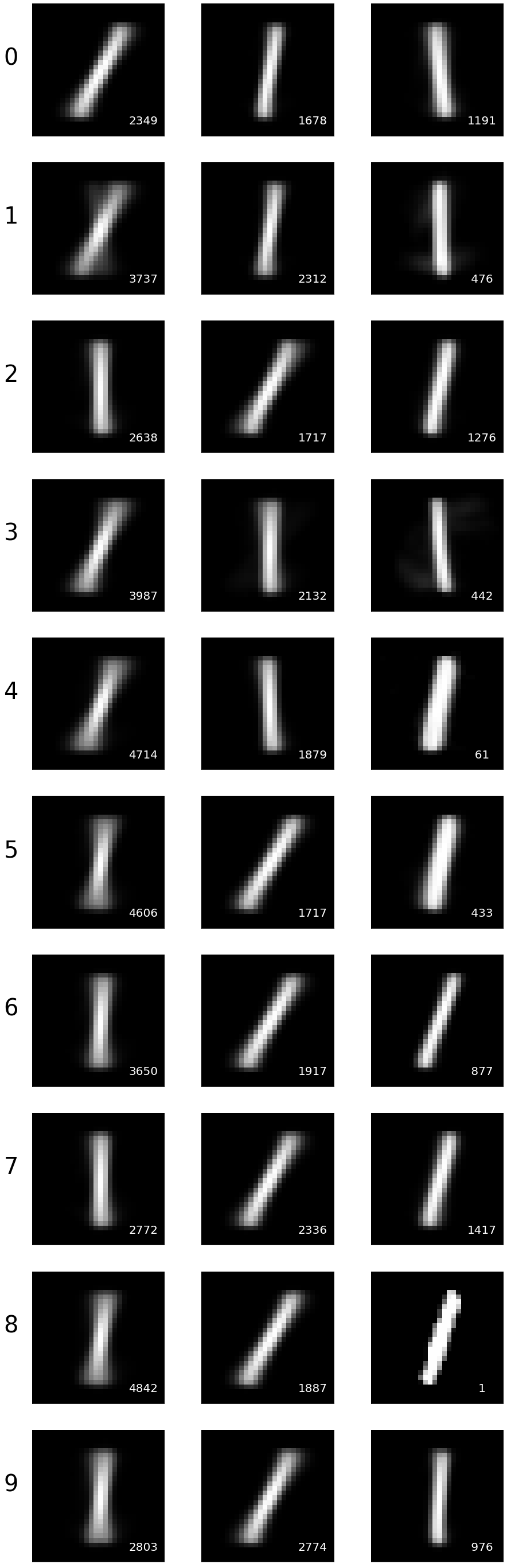}
\includegraphics[width=4cm]{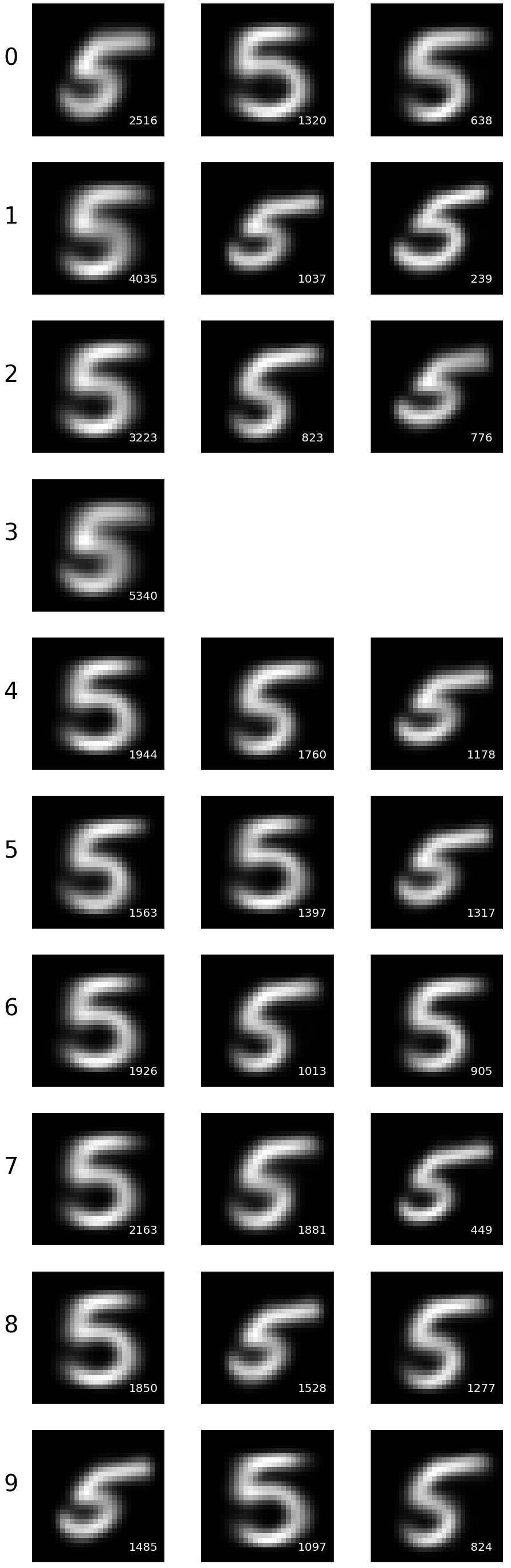}
\vspace{-5pt}
\caption{The means of the $3$ largest clusters for digits $1$ and $5$ in each of 10 trials. Trial number is shown on the left. Cluster size is shown in white on each mean image.}
\label{fig:vector_flow:random_init:mnist}
\end{figure}

Next, we evaluated the quality of subclusters for the MNIST embedding of $1$ and $5$. We ran $10$ trials of $t$-SNE with random initialization, flowed the embeddings for $40000$ iterations, and plotted the means of the limiting clusters. Fig.~\ref{fig:vector_flow:random_init:mnist} shows the means of the $3$ largest clusters for digits $1$ and $5$ in each trial. The number of subclusters found for each digit could vary substantially across trials. For example, in trial $3$ we only got a single subcluster with mostly $5$, whereas in other trials we got up to $12$ such subclusters, but it is not surprising that subcluster quality varies due to our use of random initialization and the use of a highly nonlinear flow procedure. Nevertheless, we observe that the mean images across trials are often quite similar. For example, the top two means for both digits in trial $8$ show up repeatedly in the other trials. This suggests that these subclusters identify meaningful subfeatures. 

Unlike the single Gaussians example, we cannot hope to naively average the vector fields across trials to remove artifact, since the underlying distribution for more complex datasets like MNIST may not be symmetric. In order for a similar approach to work, we would need to align embeddings generated from different runs of t-SNE. We emphasize that even for the simple Gaussian case above, different trials resulted in quite different outcomes but when we do get consistent results across trials, even if we cannot average the trials, we can get an idea of definitive substructure. Kobak et. al. provide some suggestions for how to align two distinct t-SNE embeddings \cite{transcriptomics}. This is a direction for future work.

\section{Conclusion}

We showed that the force vector in attraction-repulsion based
methods (here, in particular, tSNE) is an important additional feature. If two
points end up close to each other in the embedding but have their attractive
forces pull them in very different directions, they are probably not that similar:
they do not have the same neighbors and the neighbors that they do have
are placed in very different places in the embedding. We propose a way of
leveraging this information by flowing points along the induced vector field
until they stabilize -- you go with the flow. This leads to a naturally induced 
contraction of the clusters into subclusters and one-parameter families of 
lines respecting the structure of the data. The vector field is automatically 
generated when running t-SNE and is accessible at no extra computational 
cost. We also note that the microstructure of the vector field depends on the initialization of the embedding method and that we can arrive at a consistent view of the flowed embedding (and, hence, t-SNE) by ``averaging'' the results of several different random trials of t-SNE. We expect the idea of exploiting the force vector as an additional feature to have many other applications and implementations across a broad spectrum of attraction-repulsion based dimensionality reduction methods.

\bibliographystyle{IEEEtran}

\appendix

All code for this project was written in Python and can be found on Github: \url{https://github.com/yulanzhang/go-with-the-flow}. We generated $t$-SNE embeddings using the openTSNE package \cite{openTSNE}. We verified this implementation produces similar output to that of \href{https://scikit-learn.org/stable/modules/generated/sklearn.manifold.TSNE.html}{Scikit-learn}. Our experimental work exclusively examines $2$-dimensional embeddings. Unless we specify otherwise, we used the default settings for openTSNE. These are detailed in the online \href{https://opentsne.readthedocs.io/en/latest/index.html}{documentation} 



\begin{figure}[h!]
\centering
\includegraphics[width=4cm]{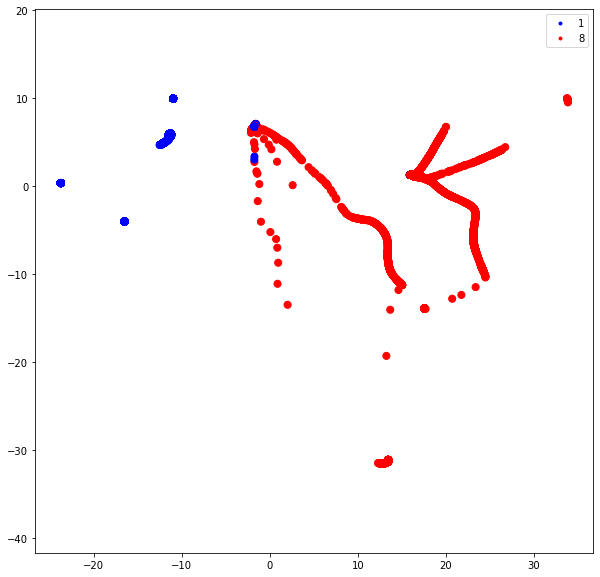}
\vspace{-5pt}
\caption{Flowed embedding from Fig.~\ref{fig:vector_flow:digits:blind} after $10000$ iterations. Blue is pants, red is purses.}
\label{fig:vector_flow:pants:10k}
\end{figure}

\begin{figure}[h]
\centering
\includegraphics[width=7cm]{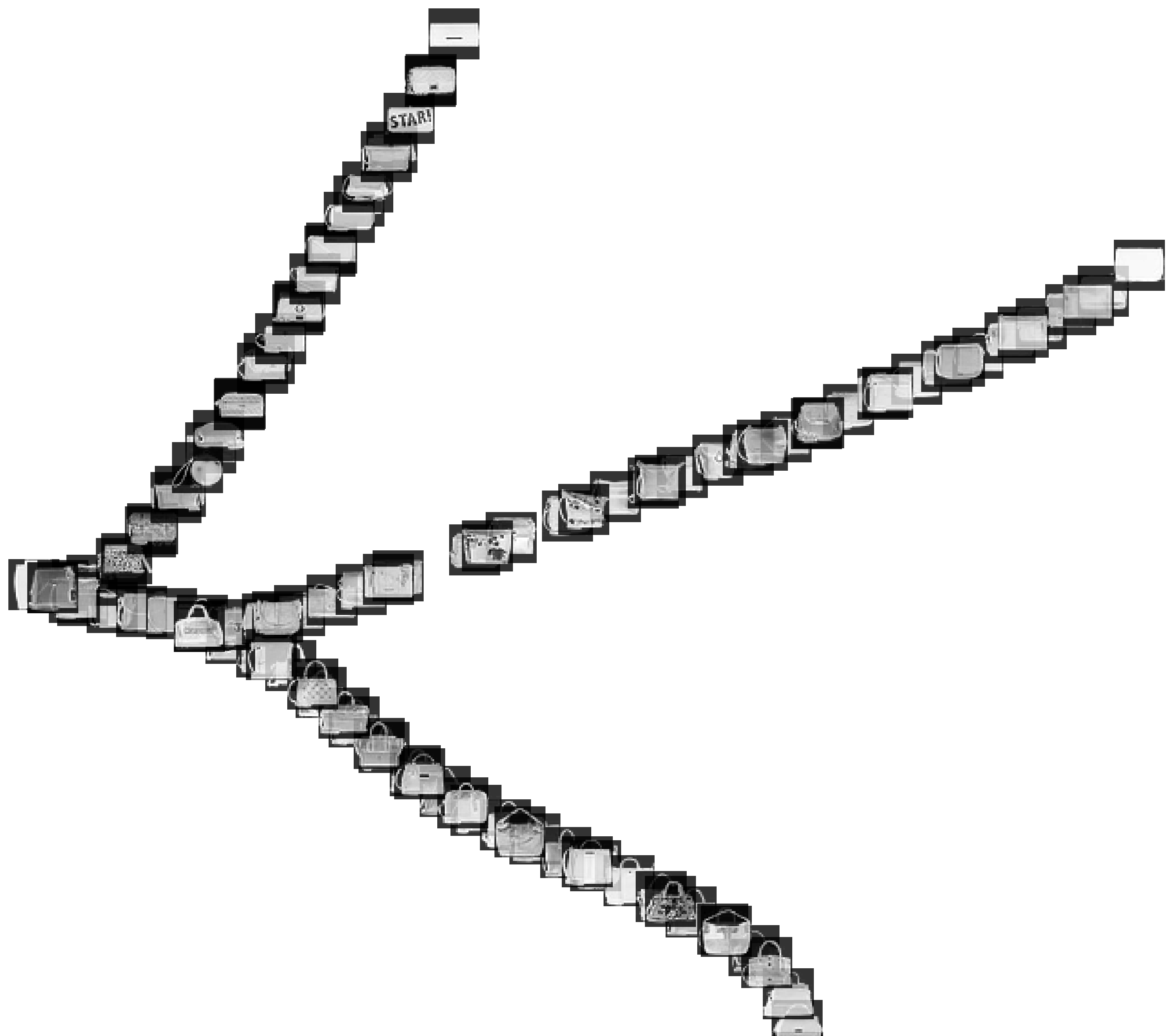}
\vspace{-5pt}
\caption{Another group of purse flow lines from Fig.~\ref{fig:vector_flow:pants:10k}. Streamlines organize the purses within the subcluster, e.g. the bottom streamline mostly contains bulky totebags or briefcases with a short, well-defined handle. The middle streamline contains shoulder bags or messenger bags where the handle and straps are less clearly visible. The top streamline shows narrow, handleless clutches.}
\label{fig:vector_flow:pants:purses}
\end{figure}

\begin{figure}[b]
\centering
\includegraphics[width=5cm]{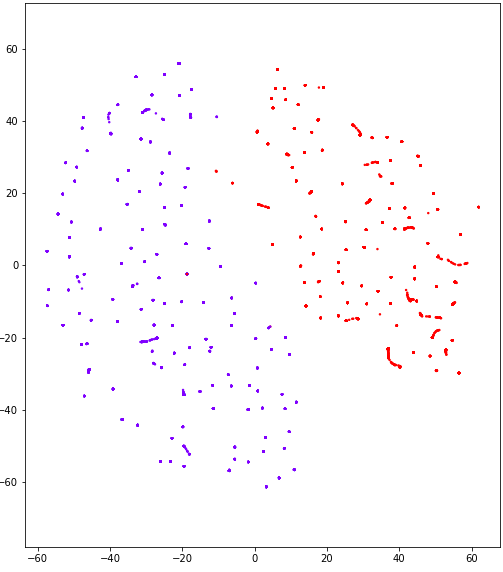}
\vspace{-5pt}
\caption{Embedding of MNIST digits $1$ and $5$, flowed along the vector field $\mathcal{F}$ induced by the original attractive t-SNE forces. Flowing along this vector field reveals excessively fine structure; almost all subclusters contain only around 50-150 samples, and images in neighboring subclusters are visually very similar. In comparison, flowing using the modified $\mathcal{\tilde F}$ in Algorithm \ref{alg:vector_flow} produces better results (Fig. \ref{fig:vector_flow:digits:blind}, \ref{fig:vector_flow:digits:40k_15}).}
\label{fig:vector_flow:orig_flow}
\end{figure}

\end{document}